\pdfoutput=1

\documentclass[11pt]{article}

\usepackage[final]{acl}

\usepackage{times}
\usepackage{latexsym}

\usepackage{url} 
\usepackage{gb4e} 
\noautomath 
\usepackage{verbatim} 
\usepackage{float} 

\usepackage{color} 

\usepackage{makecell} 

\usepackage[T1]{fontenc}

\usepackage[utf8]{inputenc}

\usepackage{inconsolata}

\usepackage{graphicx}
\usepackage{booktabs} 

\title{Generalizations across filler-gap dependencies in neural language models}

\author{
  Katherine Howitt\textsuperscript{1} \hspace{2mm}
  {\bf Sathvik Nair\textsuperscript{1, 3}} \hspace{2mm}
  {\bf Allison Dods\textsuperscript{1}} \hspace{2mm}
  {\bf Robert Melvin Hopkins\textsuperscript{1, 2}}
  \\ \textsuperscript{1}Department of Linguistics, University of Maryland
  \\ \textsuperscript{2}Department of Computer Science, University of Maryland
  \\ \textsuperscript{3}University of Maryland Institute for Advanced Computer Studies
  \\ \texttt{\{kghowitt,sathvik\}@umd.edu}
}

\begin{document}

\maketitle
\begin{abstract}
Humans develop their grammars by making structural generalizations from finite input. We ask how filler-gap dependencies, which share a structural generalization despite diverse surface forms, might arise from the input. 
We explicitly control the input to a neural language model (NLM) to uncover whether the model posits a shared representation for filler-gap dependencies. 
We show that while NLMs do have success differentiating grammatical from ungrammatical filler-gap dependencies, they rely on superficial properties of the input, rather than on a shared generalization. Our work highlights the need for specific linguistic inductive biases to model language acquisition.

\end{abstract}

\section{Introduction}
Human learners use their linguistic environment to acquire a grammar. At the same time, they come to generalizations that are not obviously signaled in the input. The central puzzle in language acquisition is to characterize the system that allows for human-like generalizations from finite input. Linguists posit that these generalizations are achieved through shared representations that allow learners to treat superficially distinct phenomena as a class \cite{chomsky_wh-movement_1977, kaplan_lexical-functional_1982, gazdar_phrase_1982, gazdar_generalized_1985, pollard_information-based_1987, postal_three_1999}. The recent success of neural language models (NLMs) has caused many to question the necessity of linguistically-specific representational systems in language learning \citep{wilcox2018rnn,wilcox_using_2023, piantadosi_modern_2023}. 

We address this renewed controversy by conducting two experiments to uncover whether NLMs posit a shared representation 
for a particular syntactic dependency: filler-gap dependencies. We consider whether an NLM recognizes filler-gap dependencies
in superficially distinct constructions, as humans do \cite{crain_rules_1985, Stowe1986ParsingWE, bever_empty_1988, traxler1996plausibility, sprouse_experimental_2016}. We further ask whether the NLM posits a shared representation for filler-gap dependencies, and thus systematically applies constraints across them.

Recent research shows NLMs can differentiate between grammatical and ungrammatical instances of filler-gap dependencies in individual constructions, but our study asks whether filler-gap dependencies are treated as a \textit{class} by the NLM. 
If a shared structural relation is learnable by an NLM, which lacks language-specific biases, then, in principle, a learner does not need to have such biases to learn that relation. Although one could learn the correct pattern through piecemeal learning of each construction individually (given enough input), a shared representation across filler-gap dependencies would allow a learner to generalize from only a subset of constructions containing filler-gap dependencies. Whether an NLM posits this shared representation is the question. 

We provide an NLM with direct evidence for a filler-gap dependency in one construction and test whether it generalizes to other constructions. In our first experiment, we augment an NLM's training data with specific instances of clefting, and in our second, topicalization. We compare performance on four constructions containing filler-gap dependencies: Wh-movement, clefting, tough-movement, and topicalization. The NLM treating filler-gap dependencies systematically would be evidence that this shared representation is learnable without language-specific inductive biases.

\section {Filler-gap dependencies}

Filler-gap dependencies share a set of properties across superficially distinct constructions, including sensitivity to islands\cite{chomsky_wh-movement_1977}. These properties persist across constructions, despite variation in semantic contribution and discourse function \cite{schutze_challenges_2015}. In psycholinguistic experiments, humans have been shown to be sensitive to gaps across filler-gap dependencies, including wh-movement \cite{crain_rules_1985, Stowe1986ParsingWE}, tough-movement \cite{bever_empty_1988}, and clefting \cite{traxler1996plausibility}, though see \citet{sprouse_experimental_2016} for variation in English relative clauses. These effects are sensitive to locality constraints, and appear to be mediated by the presence of islands \cite{phillips_real-time_2006, traxler1996plausibility,mcelree_structural_1998, omaki_hyper-active_2015}. Generalizing from surface forms on the basis of a shared representation could be critical to learning, especially if some constructions containing filler-gap dependencies do not occur frequently in the input. 

Clefting (1) is one construction that contains a filler-gap dependency. In (1a), the filler \textit{these snacks} forms a dependency with the gap site, marked with \_\_ for readability, but silent in natural language. The filler is interpreted as the object of \textit{bought}, despite not appearing linearly beside \textit{bought} in the string. Strings lacking a filler but containing a gap (1b) are ungrammatical, and when the gap is filled (i.e., an object, such as \textit{cheese}, immediately follows the verb), the acceptability pattern reverses (1c-d). In other words, neither a filler nor a gap can occur without the other. Importantly, clefts are superficially similar to sentences like (1d) which lack a filler-gap dependency, and thus are structurally quite distinct. A learner must distinguish between instances of clefting (1a) and other superficially similar sentences (1d).

\begin{exe}
\ex
  \begin{xlist}
    \ex[]{It is \textit{these snacks} that Mary bought \_\_ today.}
    \ex[*]{It is \textit{apparent} that Mary bought \_\_ today.}
    \ex[*]{It is \textit{these snacks} that Mary bought cheese today.}
    \ex[]{It is \textit{apparent} that Mary bought cheese today.}
  \end{xlist}
\end{exe}

Filler-gap dependencies occur in many constructions, including Wh-movement (2), topicalization (3), and tough-movement (4), which differ in surface form but share the filler-gap dependency and its properties. 

\begin{exe}
\ex[]{I know \textit{what} Mary bought \_\_ today.}
\ex[]{\textit{These snacks}, Mary bought \_\_ today.}
\ex[]{\textit{These snacks} are tough to buy \_\_ here.}
\end{exe}

While it might initially appear that learning could occur from simply expecting a gap when presented with a filler, properties of this dependency also include specific constraints on when they can be formed. Some structural configurations, called \textit{islands}, block the formation of a filler-gap dependency. For example, a filler-gap dependency cannot be formed inside a relative clause (e.g., \textit{that carried \_\_}) despite the fact that \textit{carried} lacks an object (i.e., is followed by a gap). The relative clause blocks the dependency, and so a gap is unacceptable regardless of the presence of a filler. Examples (5)-(8) show that all filler-gap dependencies are subject to this same restriction.  

\begin{exe}
        \ex [*]{It is \textit{these snacks} that Mary bought [the bag that carried \_\_] today.}
        \ex [*]{I know {what} Mary bought [the bag that carried \_\_] today.}
        \ex [*]{\textit{These snacks,} Mary bought [the bag that carried \_\_] today.}
        \ex [*]{\textit{These snacks} are tough to buy [the bag that carried \_\_] here.}
      \end{exe}
                        
One task for a learner is to recognize, on the basis of grammatical examples only (e.g., (1a) and (1d), but not (1b), (1c), or (5)), when each filler-gap dependency can and cannot occur. A further task is to recognize that the same properties apply to each construction containing a filler-gap dependency (2-4), and thus posit a shared representation underlying all filler-gap dependencies. 

An alternative method to generalizing would be a piecemeal learning process: learning each construction separately. For the piecemeal process to work, sufficient examples of each construction must occur in the input, and similar constraints across these constructions would arise from distinct observations. 
NLMs here provide an opportunity to test whether a shared representation can \textit{in principle} be extracted from the input without linguistic biases.

\subsection{NLMs and Filler-gap dependencies}

NLMs can learn at least some syntactic representations involving locality (see \citet{linzen2021syntactic} for a review). 
NLMs have been shown to represent shared syntactic structure across different constructions in simulated priming \citep{prasad2019using} and simulated satiation \citep{lu-etal-2024-syntactic} experiments, which compare measures from NLMs before and after exposing them to sentences with similar syntactic structures. Similarly, NLMs have been shown to generalize over syntactic structures that have been excluded from their training data \citep{jumelet-etal-2021-language,warstadt2022artificial,misra2024language,patil2024filtered}. How human-like these generalizations are is still an open question.

With respect to filler-gap dependencies, NLMs capture language-specific island constraints in English \cite{wilcox2018rnn, ozaki_how_2022, wilcox_using_2023} and Norwegian \cite{kobzeva_neural_2023} in sentences with embedded Wh-movement. \citet{ozaki_how_2022} analyze other constructions with filler-gap dependencies (clefting, topicalization, and tough-movement) and find that model performance varies by construction and is associated with the relative frequency of the constructions in texts resembling the training corpus. In other words, \citet{ozaki_how_2022} argue the model's ability to approximate human behavior is dependent on the availability of each construction type in the input. Whether this ability is modulated by a shared representation \textit{across} different constructions is not known.

Finally, \citet{lan_large_2024} investigate the extent to which NLM performance with double gap phenomena (parasitic gaps and across the board movement) is in line with human judgments. In these constructions, a gap \textit{can} occur inside an island, only if another gap is present. While they find that pretrained NLM performance is low for constructions with parasitic gaps or across the board movement, the authors show that adding examples of parasitic gaps and across the board movement to an NLM's training data adjusts its performance to be in line with human expectations, showing directly the relationship between NLM performance and surface forms in the training data. Thus, if the training data of an NLM does not contain sufficient instances of a particular construction, its ability to correctly capture the pattern of grammaticality suffers, strengthening \citet{ozaki_how_2022}'s claim that input frequency matters. 

The methodology introduced by \citet{lan_large_2024} provides a path for exploring whether NLMs make generalizations that are not apparent from simply testing a pretrained model: if the model can improve on one construction from direct training on that construction, we can ask what other effects such training might have. 
Does training a model on one construction containing a filler-gap dependency affect its performance on \textit{other} constructions containing filler-gap dependencies, the way one might expect given a shared representation?

\section{Methods}

\subsection{Measuring Filler-gap dependencies and Island Effects}
Psycholinguistic findings show structural constraints affect human expectations for gaps inside islands \citep{phillips_real-time_2006,traxler1996plausibility, Stowe1986ParsingWE}.
One way to evaluate whether an NLM's predictions align with these effects is to measure its \textit{surprisal}, the negative log probability of a word given context; less surprising words have higher probabilities. Surprisal quantifies the effect of processing difficulty \citep{levy2008expectation}. Investigating NLM surprisal at particular points in a sentence effectively treats the models like psycholinguistic subjects \citep{futrell_neural_2019}.\footnote{However, see \citet{van2021single} and \citet{huang2024large} for arguments that surprisal is not always a good estimate of human behavior for some types of syntactically complex sentences.}

To determine whether the NLMs capture syntactically relevant knowledge, we evaluate surprisal at critical regions of grammatical and ungrammatical variants of superficially similar sentences, as in (1). We compute surprisal of the region following a verb, which can either consist of a direct object (a filled gap, \textbf{-gap}) or an adverb (a gap, i.e., no direct object, \textbf{+gap}). Each string also either contains a filler (\textbf{+filler}) or does not (\textbf{-filler}). This 2x2 design is illustrated in Table \ref{tab:simple_fillereffect}, with the critical region marked in bold.
For example, the surprisal at \textit{today} in (1a) should be lower than the surprisal at \textit{cheese} in (1c) because in the latter case, given the filler \textit{these snacks}, the reader expects a gap in the object position of \textit{bought}. 
If the critical region consists of multiple words, we sum their surprisals.  

If the NLM has learned the dependency, we expect to see high surprisal in the critical regions of ungrammatical sentences: both when it encounters a gap without having seen a prior filler (1b, +gap/-filler), as well as if it has seen a filler but then encounters a filled gap (1c, -gap/+filler). Likewise, we expect low surprisal in the critical regions of grammatical sentences: if it encounters a gap after having seen a filler (1a, +gap/+filler), as well as if it sees neither filler nor gap (1d, -gap/-filler). 

To summarize these predictions, we calculate the \textbf{filler effect}: the \textit{difference} in surprisal between two sentences that are identical except for the presence of a filler \citep{wilcox2018rnn,wilcox_using_2023}. We take the surprisal for a +filler sentence and subtract the surprisal of its -filler counterpart. Based on the predictions from the previous paragraph, our filler effect predictions for \textbf{simple} (non-island) sentences are as follows: a negative filler effect in the +gap condition and a positive filler effect in the -gap condition. These predictions are in Table \ref{tab:simple_fillereffect}.

\begin{table}
  \centering
  \begin{tabular}{|>{\arraybackslash\centering}p{0.1\linewidth}|>{\centering\arraybackslash}p{0.23\linewidth}|>{\centering\arraybackslash}p{0.23\linewidth}|>{\centering\arraybackslash}p{0.17\linewidth}|} \hline 
     & \bf{+filler}& \bf{-filler}& \bf{expected effect}\\ \hline 
     \bf{+gap}& It is these snacks that Mary bought \_ \textbf{last week}.& *It is apparent that Mary bought \_ \textbf{last week}.& negative\\ \hline 
     \bf{-gap}& *It is these snacks that Mary bought \textbf{the cheese} last week.& It is apparent that Mary bought \textbf{the cheese} last week.& positive\\ \hline
  \end{tabular}
  \caption{The expected effect is the difference in the LM’s surprisal for versions of the same simple (non-island) construction with and without a filler.}
  \label{tab:simple_fillereffect}
\end{table}

The filler effect prediction for sentences with \textbf{islands} differs from the prediction for simple sentences. Filler-gap dependencies are not licensed into islands; sentences with islands are ungrammatical if they possess either a filler, a gap, or both. Only the sentences with no filler and no gap should be grammatical. Following \citet{wilcox_using_2023}, we predict an NLM with human-like performance on island effects should show filler effects around zero in sentences with islands.  
If the NLM has learned that filler-gap dependencies are always unlicensed inside an island, the presence or absence of a filler should not affect the NLM's surprisal at a gap inside an island. 
Therefore, there should be no difference between the surprisal in the +filler and -filler conditions, i.e., a filler effect of zero. These predictions are summarized in Table \ref{tab:island_fillereffect}. It is worth noting that \citet{ozaki_how_2022} have a different prediction for islands: they assume that grammaticality affects surprisal and that the NLM's surprisal will be different at the filled gap in the grammatical -gap, -filler condition. We discuss islandhood and surprisal further in Section 5.

\begin{table}
  \centering
  \begin{tabular}{|>{\centering\arraybackslash}p{0.1\linewidth}|>{\centering\arraybackslash}p{0.23\linewidth}|>{\centering\arraybackslash}p{0.23\linewidth}|>{\centering\arraybackslash}p{0.17\linewidth}|} \hline 
     & \bf{+filler}& \bf{-filler}& \bf{expected effect}\\ \hline 
     \bf{+gap}& *It is these snacks that Mary bought the bag that held \_ \textbf{last week}.& *It is apparent that Mary bought the bag that held \_ \textbf{last week}.& Closer to zero than simple effect\\ \hline 
     \bf{-gap}& *It is these snacks that Mary bought the bag that held \textbf{the cheese} last week.& It is apparent that Mary bought the bag that held \textbf{the cheese} last week.& Closer to zero than simple effect\\ \hline
  \end{tabular}
  \caption{For sentences containing islands, the expected effect is a reduction of the filler effect compared to the effect in simple sentences.}
  \label{tab:island_fillereffect}
\end{table}

\subsection{Language Model}
We estimate surprisal from a recurrent neural network (RNN) from \citet{gulordava_colorless_2018}, which is a Long-Short-Term Memory (LSTM) RNN \citep{hochreiter1997long} with two hidden layers with 650 units in each layer, trained on data from an English Wikipedia corpus (90 million tokens, or around 3 million sentences). We chose to use this model because prior research evaluating it on filler-gap dependencies has shown success in capturing human-like knowledge of filler-gap dependencies, even relative to larger models \citep{wilcox2018rnn,ozaki_how_2022,lan_large_2024,wilcox_using_2023,kobzeva_neural_2023}. Because it has transparent training data, we could carefully compare the pretrained RNN with models augmented with instances of different constructions, which we call Cleft-RNN and Topic-RNN. Details of the augmented training data for these models are explained in Section 4.
\footnote{
Transformers and LSTMs perform similarly on syntactic generalization tasks when trained on the same amounts of data, despite the transformers' lower perplexity \citep{patil2024filtered}. We did, however, replicate the results of our baseline for each construction with a pretrained GPT-2 model. See Appendix C1 for these results and more discussion on modeling choices.}

\subsection{Statistical Analysis}
We test for two effects: the first is whether the models recognize that a filler must be associated with a gap in simple sentences, and the second is whether this expectation is modulated by the presence of an island. To determine whether the RNN learned the filler-gap dependency in simple sentences, we fit a linear mixed-effects regression model following \citet{wilcox_using_2023} using surprisal as the dependent variable, sum-coded features for the presence or absence of fillers and gaps which were fixed effects. If the RNNs learn the filler-gap dependency for a particular construction, we expect to see a negative interaction term between the presence of fillers and gaps, in line with \citet{wilcox_using_2023}.

Additionally, \citet{wilcox_using_2023} fit mixed-effects models including islandhood as a fixed effect, claiming that a positive three-way interaction between the presence of fillers, gaps, and islands reflects the successful learning of island constraints. We apply this analysis, but also consider both directions of the dependency separately: unlicensed gap effects (UGE) in sentences containing a gap, and filled gap effects (FGE) in sentences without a gap  \citep{kobzeva_neural_2023}. We fit separate linear mixed-effects models for the surprisals of sentences with and without gaps, with fixed effects for fillers and islands.\footnote{Since \citet{kobzeva_neural_2023} were only testing for island effects, their regression models were fit on filler effects rather than raw surprisals based on the presence of an island. We consider the joint presence of filler and island in our analysis.} This analysis allows us to tease apart the two-way nature of the dependency and analyze whether the RNNs' failures or successes are driven by only one direction of the dependency. Success requires the regression models' coefficients to all be negative for UGEs and all be positive for FGEs for both main effects and interactions. These analyses were repeated separately for the pretrained and augmented RNNs. 
All regression models were sum-coded, included random effects for each item \citep{barr2013random}, and fit using the \texttt{Pymer} library in Python \citep{jolly2018pymer4}.
All formulas and results are reported in Appendix B.

\section{Experiments}
We evaluate an RNN's behavior on four filler-gap dependency constructions in both simple sentences and sentences with an island. 
We augment the NLM's training data with instances of a single construction and then observe the effects of that augmentation on its performance on tests of \textit{other} constructions.
In other words, we ask whether "teaching" an NLM filler-gap dependencies in one construction helps it "learn" the dependency in other constructions. If and only if the NLM acts consistently with linguists' conclusion that these constructions share an underlying representation, an improvement in performance on one construction should generalize to others.
Our implementation \footnote{\href{https://github.com/umd-psycholing/lm-syntactic-generalization}{\path{https://github.com/umd-psycholing/lm-syntactic-generalization}}} and models \footnote{\href{https://huggingface.co/sathvik-n/augmented-rnns}{\path{https://huggingface.co/sathvik-n/augmented-rnns}}} are both publicly available.

\subsection{Materials}
For embedded Wh-movement, we use the materials from \citet{wilcox_using_2023}'s experiment on complex NP islands.
For each of the other 4 constructions in (1)-(4), we test a "simple" version (no island) - where a dependency can be formed - and an island version - which does not allow for the formation of a filler-gap dependency. For topicalization, we test two versions: one ("topicalization without intro"), to closely match the materials used by \citet{ozaki_how_2022}, in which the topicalized element has no analog in the no filler sentences, and one ("topicalization with intro") in which the filler is replaced with an introductory string to control for the length of the sentence and the presence of a comma. See Tables \ref{tab:simple_fillereffect} and \ref{tab:island_fillereffect} for a schema of the design using one construction, clefting. For each construction, we generate a set of items with a fixed syntactic template and a vocabulary that varies across the set. We ensure that the lexical items are all in the RNN's vocabulary. For each item, we modulate the presence of a filler, a gap, and an island structure, generating 8 sentence types per item. Our final testing set contains 486 clefting items, 486 topicalization with intro, 161 topicalization without intro, and 243 tough-movement items. 

\subsection{Predictions}
Each graph shows the average filler effect with 95\% confidence intervals for simple and island sentences for each construction, before and after augmentation. 
For \textbf{simple} constructions, we expect to see a negative filler effect when the gap is present (blue bars) and a positive effect when the gap is absent (orange bars). This is because a human-like learner should find a gap less surprising when a filler is present than when there is no filler, and vice versa when there is no gap. (See Section 3.1 for more detail.) For \textbf{island} constructions, we expect that the confidence interval for the filler effect should overlap with zero because a human-like knowledge of islands suggests that a gap should be equally surprising regardless of the upstream presence of a filler; the same is true in the no-gap condition, using one of \citet{wilcox_using_2023}'s criteria. A less stringent relative metric for learning islands is a reduced effect relative to the filler effect in simple sentences \citep{wilcox_using_2023}; in other words, the difference in surprisal at the critical region decreases rather than disappears entirely.\footnote{For an alternative view on capturing island effects in NLMs, see Section 5 and \citet{ozaki_how_2022}}

\subsection{Experiment 1: Training on clefting}
Our initial test of the pretrained RNN yielded variation across constructions consistent with \citet{ozaki_how_2022}. Based on these results, we chose to augment the pretrained RNN's training data with instances of clefting because it fails to demonstrate knowledge of islands in clefting; also, clefting is reported as less frequent compared to Wh-movement in \citet{ozaki_how_2022}. We hypothesize that \citet{wilcox_using_2023}'s robust effects with embedded Wh-movement may be due to the relative frequency of the construction in the training corpus. 

We create a training set for clefting, using the same syntactic template for test sentences but with different lexical items.
We then retrain the RNN following the same configurations in \citet{gulordava_colorless_2018}, with training data that include all original training material and 864 additional examples of grammatical simple clefting. Half the examples contain a gap, as in (1a), half do not, as in (1d). We refer to this model as Cleft-RNN.

\begin{figure*}
  \centering
  \includegraphics[scale=0.33]{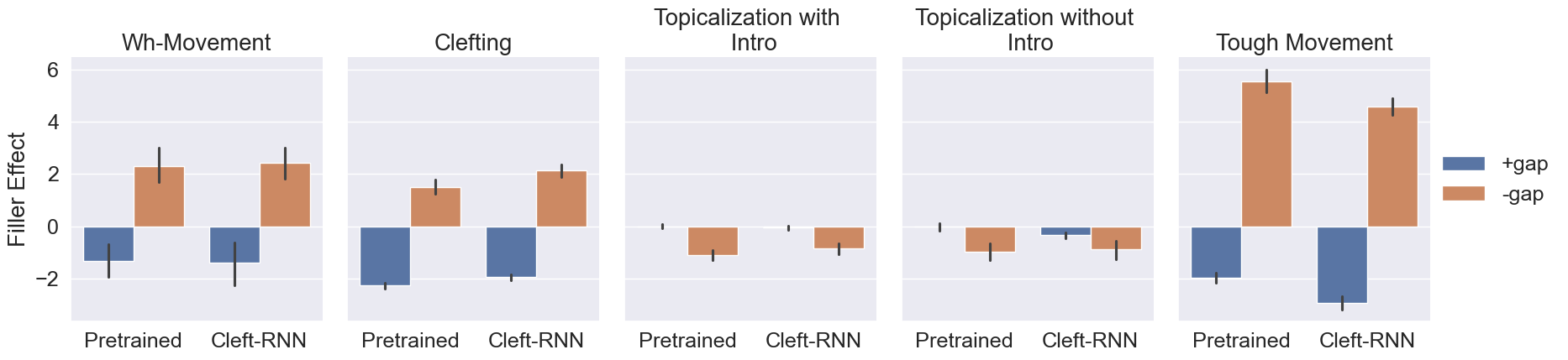}
  \caption{Filler effects for simple constructions for the pretrained model and Cleft-RNN.}
  \label{fig:cleft_rnn}
\end{figure*}

\begin{figure}[ht]
  \vspace{-2ex}
  \centering
  \includegraphics[scale=0.27]{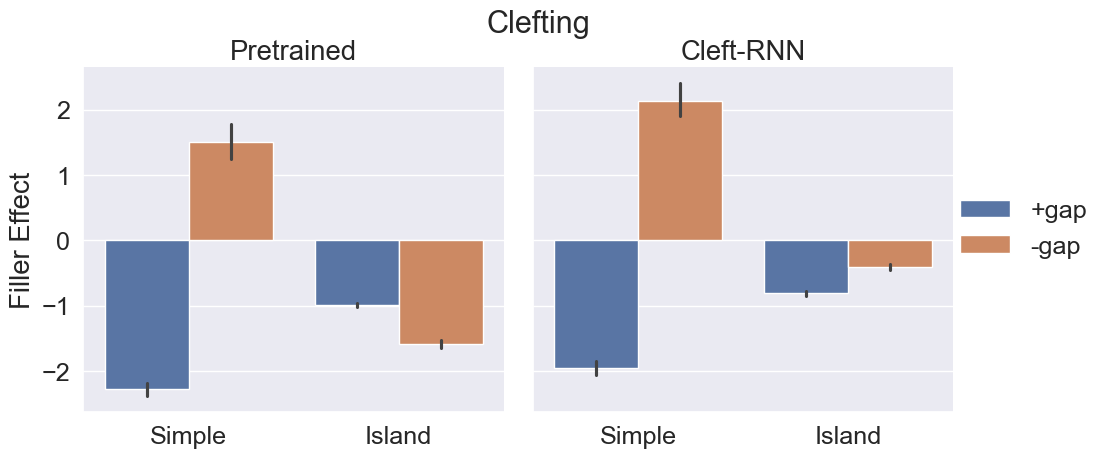}
    \includegraphics[scale=0.27]{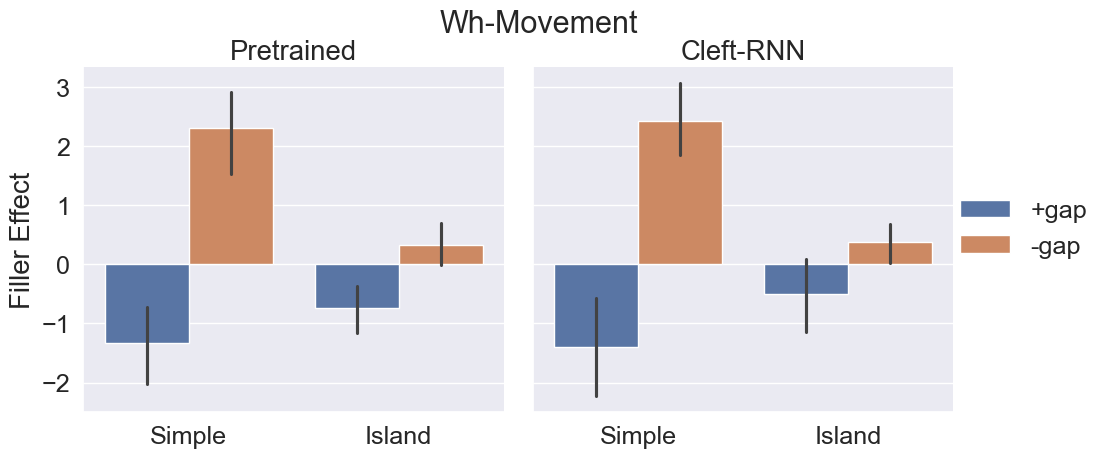}
  \includegraphics[scale=0.27]{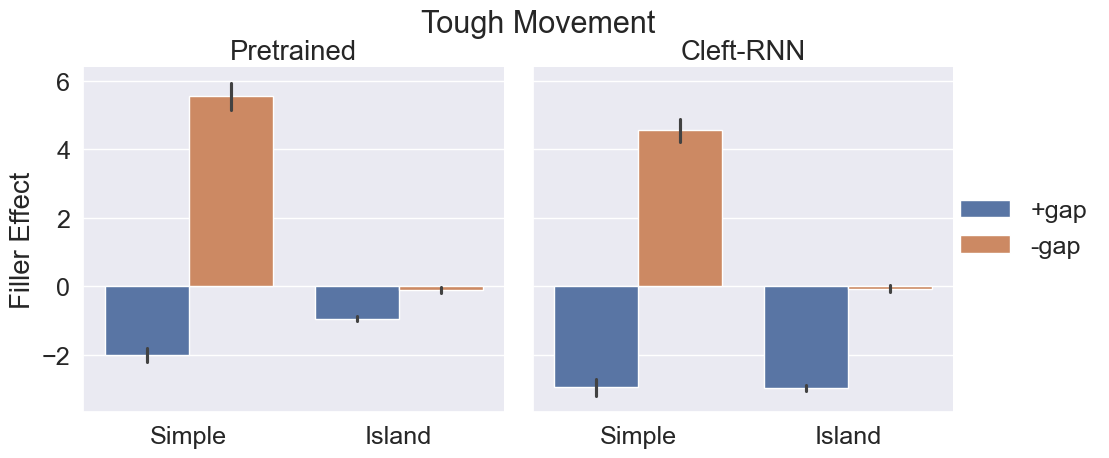}

  \caption{Filler effects for simple and island constructions for the pretrained model and Cleft-RNN. Since this dependency was not learned for topicalization, we do not display these results.}
  \label{fig:cleft_rnn_islands}
\end{figure}

\subsubsection{Results}
\textit{Simple constructions.} We first present the filler effects for the simple sentences of each construction of the pretrained RNN (before augmentation) and Cleft-RNN, plotted in Figure \ref{fig:cleft_rnn}. Testing the pretrained RNN on simple constructions, we replicate \citet{wilcox_using_2023}'s findings for embedded Wh-movement. The pretrained RNN also shows the desired filler effect pattern in simple sentences for clefting and tough-movement, but not for either form of topicalization. For each construction type, we looked at the two-way interaction of filler and gap, confirming a positive result for Wh-movement, clefting, and tough-movement (negative interaction terms with $p < 0.001$). The interaction effects for topicalization were positive, indicating that the pretrained RNN did not learn the dependency in either type of topicalization constructions.

Training on clefting had no significant effect on knowledge of the dependency in simple sentences of any construction, confirmed both by the qualitative appearance of the graphs and by the mixed-effects models for each construction type (negative interaction terms with $p < 0.001$). \footnote{We did observe changes in effect size, but because we consider knowledge of filler-gap dependencies to be binary (either learned or not learned), it is difficult to draw conclusions from minor changes in the effect size that do not change the status of the significant interaction.} Since Cleft-RNN did not learn the dependency in either form of topicalization, we do not report island effects. 

\textit{Island constructions.} The results for constructions containing islands before and after augmentation, presented in Figure \ref{fig:cleft_rnn_islands}, are less straightforward. For no construction did the pretrained RNN meet the most stringent criteria for recognizing island constraints: that is, both filler effects in the island condition equaling zero. We do see varying degrees of \textit{reduction} in the filler effects for each construction in the island vis-a-vis the simple condition. We consider each direction of the dependency separately: filled gap effects (FGE, orange) and unlicensed gap effects (UGE, blue).

The mixed-effects model using \citet{wilcox_using_2023}'s methods shows negative filler-gap interaction terms and positive three-way interaction terms for Wh-movement, clefting, and tough-movement, suggesting that the pretrained RNN correctly captures island constraints for all constructions where it knows the simple dependency. However, this result is at odds with our qualitative findings, which suggest the model did not learn the relevant generalization in clefting. We observe that in the -gap condition (orange bars), the filler effect is equal in magnitude in both the simple and island conditions. In fact, the filler effect is negative, suggesting high surprisal in the grammatical (-filler, -gap) sentence (see Table \ref{tab:island_fillereffect}). This qualitative result is confirmed by our separate mixed-effects model for FGEs, which shows statistically significant positive coefficients for Wh-movement and tough-movement $(p < 0.001)$, but not for clefting. In the regression models for UGE, we observe statistically significant negative coefficients for all three constructions $(p < 0.001)$. In other words, the pretrained RNN is not sensitive to FGEs for clefting constructions, consistent with the qualitative pattern.

We now review Cleft-RNN's behavior with island constructions. The direction and magnitude of the effects in the three-way interaction model are consistent with the conclusion that Cleft-RNN captures island constraints in clefting and Wh-movement. For the clefting construction (in other words, when presented with examples structurally identical to those it was trained on), our regression models for FGEs and UGEs support this result. All coefficients for the UGEs are negative, and all coefficients for the FGEs are positive. Cleft-RNN is less sensitive to the presence of an upstream filler at a filled gap in an island construction than the pretrained RNN, though qualitatively the grammatical form is still more surprising than the ungrammatical form. For Wh-movement, Cleft-RNN's confidence interval of the filler effect for islands in the +gap condition overlaps with zero, achieving our most stringent criterion for displaying knowledge of island constraints. Results were statistically significant, both for clefting $(p < 0.001)$ and for Wh-movement $(p < 0.01)$, which was tested on a smaller stimulus set. 

In tough-movement, however, augmentation has a detrimental effect. The magnitude of the filler effect in the +gap condition for islands is equivalent to that in the simple cases. The positive, non-significant interaction term for the regression model for UGEs in tough-movement supports this observation; Cleft-RNN lacks sensitivity to islands in tough-movement constructions.

For topicalization, our regression models do not have the correct signs for either construction type, confirming, as a qualitative inspection of Figure \ref{fig:cleft_rnn} suggests, that Cleft-RNN does not learn the dependency in these constructions. 

\subsection{Experiment 2: Training on topicalization}

Since neither the pretrained RNN nor Cleft-RNN are able to arrive at the correct generalization for cases of topicalization, we now determine if providing the pretrained RNN with positive direct evidence of topicalization is sufficient for learning this dependency.
We follow a similar procedure to the previous experiment, generating sentences from the same syntactic template for topicalization with intro and ensuring the lexical items do not appear in the testing sentences.
We augment the RNN's Wikipedia corpus with 864 grammatical examples of topicalization, half with a gap and half without a gap, and train it using the hyperparameters in \citet{gulordava_colorless_2018}. The augmented model is referred to here as Topic-RNN.

\subsubsection{Results}
\begin{figure}
  \centering
  \includegraphics[scale = 0.28]{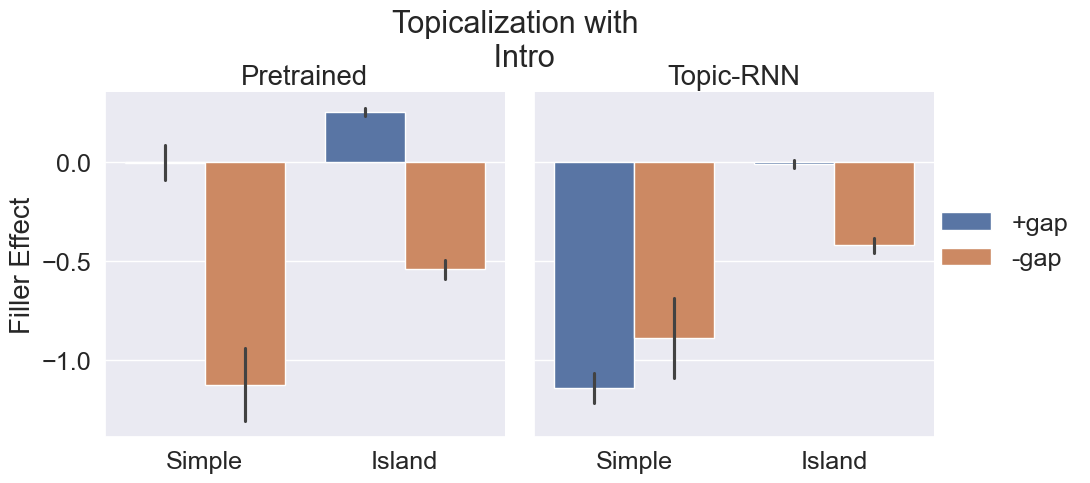}
    \includegraphics[scale = 0.28]{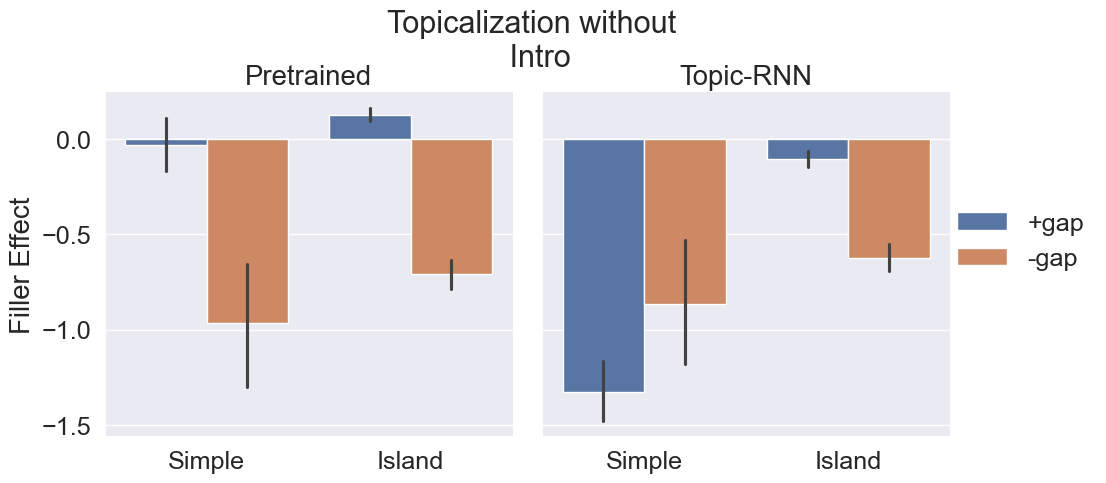}
  \caption{Filler effects for simple and island constructions for the pretrained model and Topic-RNN.}
  \label{fig:topic_rnn}
\end{figure}

Figure \ref{fig:topic_rnn} shows filler effects in the pretrained RNN and Topic-RNN. Training explicitly on simple instances of topicalization does not lead the model to posit the dependency in both directions. Here, positive evidence is not enough to learn even the simple dependency. 
The regression model does not show significant effects for basic filler-gap licensing in Topic-RNN.

Topic-RNN \textit{does} learn that the presence or absence of an upstream filler should modulate surprisal at a gap (UGEs, blue bars), but it fails to learn the correct relationship between a filled gap and the absence of an upstream filler (FGEs, orange bars). In fact, Topic-RNN's surprisal is consistent with the non-human-like hypothesis that \textit{the cheese} in the sentence \textit{The snacks, Mary bought the cheese last week} is less surprising than in the sentence \textit{In fact, Mary bought the cheese last week}. 

\section{Discussion}
In this paper, we test whether NLMs can generalize knowledge of filler-gap dependencies across different constructions when their input is augmented with only one construction containing a filler-gap dependency. The pattern of knowledge of the pretrained RNN potentially reflects piecemeal learning, where the frequency of particular constructions modulates the model's recognition of the filler-gap dependency for each construction type individually \cite{ozaki_how_2022}. However, based on the pretrained results alone
we cannot determine whether the NLM's inferences for one construction are based on others, hence the need for an augmentation-based procedure.
Experiment 1 found that while Cleft-RNN behaves differently than the pretrained RNN on clefting, Wh-movement, and tough-movement, it fails to generalize systematically across all the types of filler-gap dependencies we test. Cleft-RNN's failure to learn the relevant dependency for simple topicalization sentences further confirms that these models do not arrive at their knowledge of filler-gap dependencies through a shared representation.

Cleft-RNN does improve its representation of island constraints in clefting, the construction it was augmented with. However, this improvement still preserves the incorrect prediction for grammaticality. In this case, positive evidence of grammatical forms is still insufficient for human-like learning. This finding supports a conclusion drawn by \citet{lan_large_2024}: that given sufficient evidence of a construction, NLMs can arrive at a correct representation of the constraints. However, the ability to generalize from one construction appears weak at best.

Cleft-RNN's filler effect for Wh-movement is the only instance among our findings where an NLM achieves the most stringent measure of islands: a confidence interval overlapping with zero. However, we are cautious to over-interpret this finding: our test set for Wh-movement was far smaller than that of the other construction types.

Further, the failure of Cleft-RNN to capture islands in tough-movement relative to the pretrained RNN highlights a more pressing issue with NLMs that rely only on surface distributions: exposure to one construction type can cause a degradation in an NLM's knowledge of a different type, when the two share superficially similar characteristics at odds with the dependency. The learner would then need even more positive direct evidence of the other type to offset such erroneous conclusions. 

We found that the NLM we tested fares worse at recognizing island constraints than past studies would suggest: both the pretrained RNN and our augmented models failed to arrive at the most stringent measure of islandhood in all cases but Wh-movement in Cleft-RNN. We suspect this is greatly influenced by the frequency of surface forms of particular constructions, as hypothesized by \citet{ozaki_how_2022}.

Why, then, did \citet{wilcox_using_2023}'s method of using the presence of fillers, gaps, \textit{and} islands as predictors of surprisal yield a significant interaction for islandhood in instances where the filler effect suggested otherwise? We believe that the interaction collapses effects across different combinations of features. Our NLMs succeed with UGEs, which likely obscures their corresponding failure to recognize FGEs.
However, the failure in FGEs suggests that the NLM is recognizing neither grammaticality nor the presence of an island. For FGEs, the filler effect is in the wrong direction; grammatical continuations (i.e., those without a filler or a gap) are more surprising than ungrammatical ones (a filler and no gap). This contradicts both the measures for islands proposed by \citet{wilcox_using_2023} and \citet{ozaki_how_2022}, who suggest that rather than no difference at the gap site, surprisal should align with grammaticality. Here we find that surprisal does not align with either measure and is in fact showing the reverse pattern for grammaticality.


The results of Experiment 1 strongly suggest that the NLM arrives at its knowledge of filler-gap dependencies through piecemeal learning and that positive direct evidence of a filler-gap dependency in each construction is required to learn the dependency for that construction. We conducted Experiment 2 to explore whether positive evidence of simple topicalization sentences is sufficient for Topic-RNN to make predictions consistent with a human-like understanding of both the simple dependency and islands. Topic-RNN learns to expect a gap given a filler, but fails to learn the other direction of the dependency: that in the absence of a filler, there should be no gap. The model's failure to learn the simple topicalization dependency even in the face of direct evidence is an additional challenge to claims that language-specific biases are not necessary to learn such dependencies.

Taken together, the results from Experiments 1 and 2 show that NLMs do not generalize from a shared representation to learn filler-gap dependencies. Instead, they rely heavily on input that closely aligns with individual constructions. Further, in cases such as topicalization, NLMs appear to struggle with learning the dependency. Our findings are particularly important as researchers consider in what ways NLMs might and might not serve as good proxies for language learners. Our work reiterates the importance of  specific linguistic inductive biases to model language acquisition. 

\section*{Acknowledgements}
We appreciate feedback from Jeff Lidz, Colin Phillips, Philip Resnik, Naomi Feldman, Hal Daumé III, Bill Idsardi and other members of the UMD Psycholinguistics Workshop, Computational Cognitive Science group, and CLIP labs. 
Jiayi Lu, Nur Lan, and Suhas Arehalli provided insightful suggestions regarding our methodology. 
Sathvik Nair was supported by the ONR MURI Award N00014-18-1-2670 and NSF GRFP Grant No. DGE 2236417.

\bibliography{custom}

\clearpage
\newpage

\appendix
\section{RNN Training Configuration}
\label{sec:hparams}
We trained our RNN models using the hyperparameters of the model that achieved the lowest perplexity on the validation set in \citet{gulordava_colorless_2018}.
Test set perplexities were 0.45 for the model augmented with clefting and 50.34 for the model augmented with topicalization.
The pretrained RNN model had a perplexity of 51.9; a slight decrease in perplexity is expected with the additional grammatical input.
\begin{table}[ht]
    \centering
    \begin{tabular}{|c|c|}
    \hline
        Embedding Size & 650 \\
        Hidden Units & 650\\
        Layers & 2\\
        Dropout & 0.2\\
        Batch Size & 128\\
        Learning Rate & 20\\
        Epochs & 40\\
    \hline
    \end{tabular}
    \caption{Training parameters used by the best performing LSTM in \citet{gulordava_colorless_2018}.}
    \label{tab:grnn_params}
\end{table}

\section{Results of Statistical Tests}
\label{sec:appendix}
\subsection{Basic Filler-Gap Licensing}
\label{sec:basic_stats}
We first fit linear mixed-effects regression models testing for the filler-gap interaction in the simple sentences alone, with the formula \texttt{surprisal $\sim$ wh * gap + (1 | item)}, following \citet{wilcox_using_2023}.
In Table \ref{tab:basic_fg}, we report the value of the filler-gap interaction term.
For all negative filler-gap interactions, $p < 0.001$, except for Wh-movement in Cleft-RNN and Topic-RNN ($p < 0.01)$.
Although we do report negative filler-gap interactions for Topic-RNN, these are not statistically significant.
\begin{table}[ht]
  \centering
\begin{tabular}{|p{0.8in}|p{0.6in}|p{0.5in}|p{0.5in}|}
  \hline
  Construction & Pretrained RNN & Cleft-RNN & Topic-RNN \\
  \hline
  Wh-Movement & -0.905 & -0.954 & -1.128 \\
  Clefting & -0.947 & -1.024  &  -0.610	\\
  Topicalization without Intro & 0.136 & 0.545 & -0.063 \\
  Topicalization with Intro & 0.200 & 0.8 & -0.115 \\
  Tough-Movement & -1.887 & -1.878 & -1.871 \\
  \hline
  \end{tabular}
  \caption{Filler-Gap interaction effects for Simple sentences}
  \label{tab:basic_fg}
\end{table}

\subsection{Island Effects}
\label{sec:island_stats}
We then fit separate linear mixed-effects regression models to determine if the presence of islands had an effect on model surprisal.
We tested for the effect of islands separately in sentences without and with gaps (Filled Gap Effects and Unlicensed Gap Effects, respectively, noted as FGE and UGE for short), using the formula \texttt{surprisal $\sim$ filler * island + (1 | item)}. Levels of significance are reported as follows: . $p < 0.1$** $p < 0.01$ *** $p < 0.001$
In Table \ref{tab:cleft_rnn_islands}, we report the effects for Experiment 1.
We test for island effects in the constructions that show a basic filler-gap licensing effect (clefting, Wh-movement, and tough-movement), showing main effects and interactions for surprisal from Cleft-RNN and the pretrained RNN.
Table \ref{tab:topicalization_stat} shows the effects for Experiment 2 on both types of topicalization constructions for Topic-RNN and the pretrained RNN.
\begin{table}
    \centering
      Clefting FGE, Pretrained RNN

    \begin{tabular}{p{0.65in}|p{0.5in}|p{0.35in}|p{0.7in}}
     &  Estimate &    SE &  T-stat  \\ \hline 
     filler &    -0.039 & 0.058 &  -0.669 \\       island &     2.869 & 0.076 &  37.653*** \\ filler:island &     3.088 & 0.116 &  26.712*** \\ \hline 
     \end{tabular} \\
      \vspace{1em} Clefting UGE, Pretrained RNN

     \begin{tabular}{p{0.65in}|p{0.5in}|p{0.35in}|p{0.7in}}
     \hline filler &    -1.636 & 0.030 & -54.330*** \\       island &    -1.456 & 0.040 & -36.521*** \\ filler:island &    -1.293 & 0.060 & -21.469*** \\ 
     \hline
    \end{tabular}    \vspace{1.5em}

    Clefting FGE, Cleft-RNN\\
\begin{tabular}
{p{0.65in}|p{0.5in}|p{0.35in}|p{0.7in}}
 \hline filler & 0.866 & 0.054 & 16.158*** \\ island & 3.359 & 0.069 & 48.508*** \\  filler:island & 2.547 & 0.107 & 23.750*** \\ \hline
 \end{tabular}\\
 \vspace{1em} Clefting UGE, Cleft-RNN
 \begin{tabular}{p{0.65in}|p{0.5in}|p{0.35in}|p{0.7in}}
 \hline filler & -1.382 & 0.031 & -44.183*** \\ island & -0.827 & 0.041 & -19.973***\\ filler:island & -1.148 & 0.063 & -18.341*** \\ \hline
 \end{tabular}

\vspace{1.5em}     Wh-Movement FGE, Pretrained RNN

\begin{tabular}{p{0.65in}|p{0.5in}|p{0.35in}|p{0.7in}}
\hline     filler &     1.310 & 0.333 &  3.932*** \\      island &     3.350 & 0.333 & 10.055*** \\
filler:island &     1.976 & 0.666 &   2.965** \\ \hline
\end{tabular} \\ \vspace{1em}

Wh-Movement UGE, Pretrained RNN \\
\begin{tabular}{p{0.65in}|p{0.5in}|p{0.35in}|p{0.7in}}
\hline
filler &    -1.030 & 0.399 &   -2.582* \\  island &    -0.158 & 0.399 &    -0.395 \\ filler:island &    -0.587 & 0.798 &    -0.737 \\ \hline 
\end{tabular}
\\ \vspace{1.5em}
Wh-Movement FGE, Cleft-RNN \\
\begin{tabular}{p{0.65in}|p{0.5in}|p{0.35in}|p{0.7in}}
\hline
filler & 1.392 & 0.386 & 3.604*** \\
island & 2.586 & 0.386 & 6.696*** \\
filler:island & 2.047 & 0.772 & 2.65** \\ \hline
\end{tabular} 
\\ \vspace{1em}
Wh-Movement UGE, Cleft-RNN \\ 
\begin{tabular}{p{0.65in}|p{0.5in}|p{0.35in}|p{0.7in}} \hline
(Intercept) & 21.403 & 0.725 & 29.525*** \\
filler & -0.953 & 0.400 & -2.385* \\ island & 0.751 & 0.400 & 1.881 . \\ filler:island & -0.895 & 0.799 & -1.12 \\ \hline 
\end{tabular}
\\ \vspace{1.5em}
Tough-Movement FGE, Pretrained RNN
\begin{tabular}{p{0.65in}|p{0.5in}|p{0.35in}|p{0.7in}} 
\hline
filler &     2.729 & 0.093 &    29.3*** \\   island &     3.858 & 0.122 &  31.637*** \\
filler:island &     5.655 & 0.186 &  30.358*** \\ \hline
\end{tabular}
\\ \vspace{1em}
 Tough-Movement UGE, Pretrained RNN
\begin{tabular} {p{0.65in}|p{0.5in}|p{0.35in}|p{0.7in}} \hline
filler &    -1.468 & 0.060 & -24.309*** \\   island &    -0.425 & 0.080 &  -5.338*** \\ filler:island &    -1.046 & 0.121 &  -8.659*** \\\hline
\end{tabular} \\ \vspace{100em} 
\end{table}

\begin{table}[ht]
    \centering
Tough-Movement FGE, Cleft-RNN
\begin{tabular}{p{0.65in}|p{0.5in}|p{0.35in}|p{0.7in}} & Estimate & SE & T-stat 
\\ \hline  filler & 2.258 & 0.084 & 26.814*** \\ island & 3.209 & 0.108 & 29.618*** \\filler:island & 4.634 & 0.168 & 27.512*** \\ \end{tabular}
\\ \vspace{1em}

Tough-Movement UGE, Cleft-RNN \\

\begin{tabular} {p{0.65in}|p{0.5in}|p{0.35in}|p{0.7in}} \hline  filler & -2.943 & 0.076 &  -38.892*** \\ island & -1.183 & 0.100 & -11.865*** \\ filler:island & 0.012 & 0.151 & 0.076 \\\hline
\end{tabular}
\caption{Regression Model Coefficients for FGEs and UGEs based on the pretrained RNN and Cleft-RNN for Clefting, Wh-Movement, and tough-movement constructions.}
    \label{tab:cleft_rnn_islands}
\end{table}

\begin{table}
    \centering
With Intro FGE, Pretrained RNN
\begin{tabular}{p{0.65in}|p{0.5in}|p{0.35in}|p{0.7in}}
&  Estimate &    SE &     T-stat \\ \hline
filler &    -0.832 & 0.050 & -16.733*** \\
island &     0.899 & 0.066 &   13.63*** \\
filler:island &    -0.582 & 0.099 &  -5.851*** \\
\hline
\end{tabular}
\\ \vspace{1em}

With Intro UGE, Pretrained RNN
\begin{tabular}{p{0.65in}|p{0.5in}|p{0.35in}|p{0.7in}} \hline
     filler &     0.125 & 0.025 &   4.945*** \\      island &    -1.049 & 0.034 & -31.221*** \\ filler:island &    -0.257 & 0.051 &   -5.09*** \\ \hline
\end{tabular}
\\ \vspace{1.5em}

With Intro FGE, Topic-RNN
\begin{tabular}{p{0.65in}|p{0.5in}|p{0.35in}|p{0.7in}}
\hline
filler & -0.653 & 0.045 & -14.628*** \\
island & 1.718 & 0.058 & 29.486*** \\
filler:island & -0.467 & 0.089 & -5.238*** \\ \hline 
\end{tabular}
\\ \vspace{1em}

With Intro UGE, Topic-RNN
\begin{tabular}{p{0.65in}|p{0.5in}|p{0.35in}|p{0.7in}} \hline
filler & -0.574 & 0.024 & -24.335*** \\
island & -1.391 & 0.031 & -44.272*** \\
filler:island & -1.130 & 0.047 & -23.961*** 
\end{tabular}
\\ \vspace{1.5em}

With Intro FGE, Pretrained RNN
\begin{tabular}{p{0.65in}|p{0.5in}|p{0.35in}|p{0.7in}}
\hline
 (Intercept) &    21.855 & 0.129 & 170.019*** \\
 filler &    -0.838 & 0.086 &  -9.741*** \\
 island &     0.952 & 0.114 &   8.345*** \\
 filler:island &    -0.256 & 0.172 &     -1.487 \\ \hline  
 \end{tabular}
\\ \vspace{1em}

Without Intro UGE, Pretrained RNN
\begin{tabular}{p{0.65in}|p{0.5in}|p{0.35in}|p{0.7in}} \hline
 (Intercept) &    17.170 & 0.077 & 223.556*** \\       filler &     0.048 & 0.040 &      1.197 \\       island &    -0.831 & 0.054 & -15.394*** \\
 filler:island &    -0.157 & 0.081 &    -1.938 . 
\\ \hline \end{tabular}
\\ \vspace{1.5em}

Without Intro FGE, Topic-RNN
\begin{tabular}{p{0.65in}|p{0.5in}|p{0.35in}|p{0.7in}}
\hline
filler & -0.746 & 0.077 & -9.647*** \\
island & 1.629 & 0.101 & 16.114*** \\
filler:island & -0.245 & 0.155 & -1.582 \\ \hline
\end{tabular}
\\ \vspace{1em}
Without Intro UGE, Topic-RNN
\begin{tabular}{p{0.65in}|p{0.5in}|p{0.35in}|p{0.7in}} \hline
filler & -0.718 & 0.044 & -16.33*** \\ island & -1.399 & 0.059 & -23.89*** \\ 
filler:island & -1.222 & 0.088 & -13.905*** \end{tabular}\

\caption{Regression Model Coefficients for FGEs and UGEs in Topicalization}
\label{tab:topicalization_stat}
\end{table}

Using the method of \citet{wilcox_using_2023} to test for island effects, we trained linear mixed-effects regression models for the surprisals under each NLM for each construction using the formula \texttt{surprisal $\sim$ wh * gap * island + (1||item)}.
All effect sizes we report are statistically significant $(p < 0.001)$, except for the pretrained RNN on Wh-islands $(p < 0.05)$, likely due to the comparatively smaller size of the dataset, and on topicalization. 
For the model trained on instances of clefting, where the only non-significant effects are reported on topicalization.
For the model that is trained on instances of topicalization, the filler-gap effects are are significant($p < 0.05$) rather than $(p < 0.001)$, suggesting that the effect was not sufficient to make human-like generalizations.
According to \citet{wilcox_using_2023}, the model has arrived at the appropriate generalization if the filler-gap interaction is negative, but the three-way interaction (listed under \textit{Island} in the table) is positive. 
This, however, must be corroborated by the qualitative results with the filler effects.

\section{Limitations}

\subsection{Modeling}
\begin{figure*}[b]
    \vspace{-18cm}
    \centering
    \includegraphics[scale=0.33]{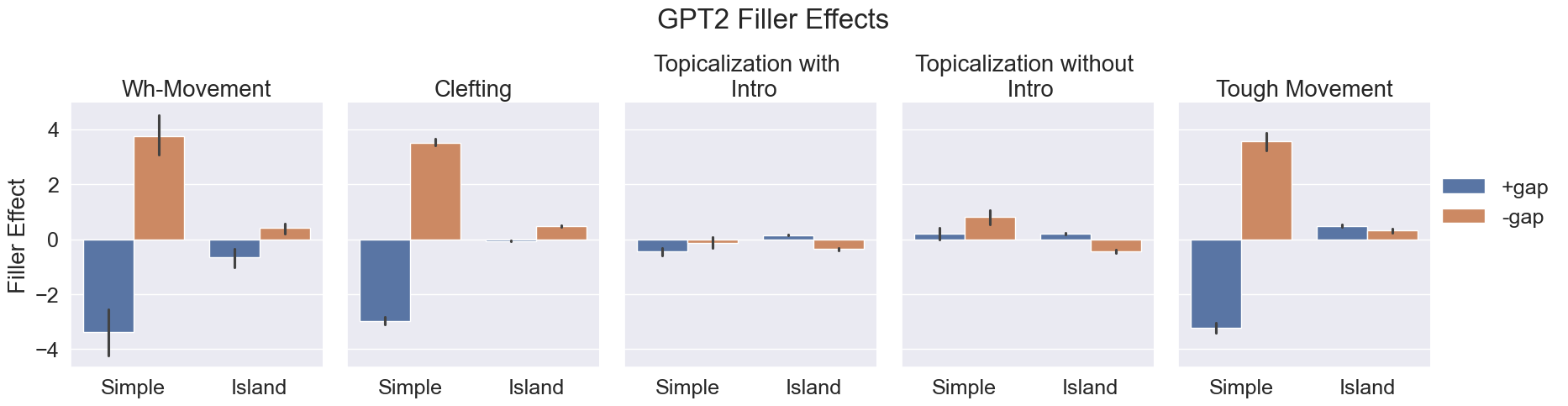}
    \caption{Filler effects for a pretrained GPT-2 model, evaluated on the same set of sentences as the RNNs.}
    \label{gpt}
\end{figure*}

Although Transformer models like GPT2 are used more often, we decided to work with an RNN specifically to be consistent with the existing literature on NLMs and filler-gap dependencies, and since retraining a GPT2-scale model was not feasible with the computing resources we had available. However, the pretrained GPT2-small model \citep{radford2019language} largely produced the same qualitative patterns as the RNN. (Figure \ref{gpt})

In future work, we aim to train a smaller Transformer model on the same corpus as the RNN to determine if architecture has any effects when training data is kept constant, but see recent work from \citet{patil2024filtered} that shows that LSTMs and Transformer models with comparable size perform equally well on measures of linguistic generalization.
That said, however, results from \citet{murty-etal-2023-grokking} show that increasing the training time for small transformer LMs does improve syntactic generalization on out-of-domain data; future work should determine if this ability extends to filler-gap dependencies.
Finally, we did not fine tune the model with additional examples since we wanted to simulate learning a richer corpus, instead of determining how the model's weights would change \textit{after} being provided explicit examples of the constructions.

\subsection{Training Data}
The Wikipedia corpus the RNN was trained on contains mostly grammatical utterances and is far less noisy than the data a child is exposed to.
However, since the model failed to extract the generalization across constructions with consistently well-formed input, more varied input would likely show a similar effect, if not worse.
It is possible that augmenting the model with more examples of explicit constructions may improve its performance, but this is not likely to change the observed effects significantly \citep{van2019quantity}.

\begin{table*}[t]
\vspace{-7cm}
  \centering
\begin{tabular}{|p{0.8in}|c|c|c|c|c|c|}
  \hline
  Construction  & \multicolumn{2}{c|}{Pretrained RNN} & \multicolumn{2}{p{0.5in}|}{Cleft-RNN} & \multicolumn{2}{p{0.5in}|}{Topic-RNN} \\
  \hline
    & Filler-Gap & Island & Filler-Gap & Island & Filler-Gap & Island \\
  \hline
  Wh-Questions & -3.621 & 2.563 & -3.816 & 2.942 & -4.513 & 3.04\\
  Clefting & -3.788 & 4.381 & -4.09 & 3.695 & -2.439 & 2.468\\
  Topicalization without Intro & 0.936 & -0.099 & 0.545 & -0.112 & -0.461 & 0.977 \\
  Topicalization with Intro & 1.120 & -0.325 & 0.8 & -0.437 & -0.252 & 0.662  \\
  Tough-Movement & -6.773 & 6.954 & -7.513 & 4.622 & -7.483 & 5.321 \\
  \hline
  \end{tabular}
  \caption{Interaction effects from mixed-effects regression models.}
  \label{tab:effect_sizes}
\end{table*}

\end{document}